\documentclass{svproc}
\usepackage{amsmath,amssymb,amsfonts}
\usepackage{indentfirst}
\usepackage{graphicx}
\graphicspath{ {./fig/} }
\usepackage{bibentry}
\usepackage[backend=biber, style=ieee, url=false]{biblatex}
\usepackage{algorithm}
\usepackage{algorithmic}
\usepackage{booktabs}

\usepackage{url}

\begin{document}
\mainmatter              
\title{Automated Materials Spectroscopy Analysis using Genetic Algorithms}
%
%
\author{Miu Lun Lau \inst{1} \and Min Long \inst{1} \and Jeff Terry \inst{2}}
%
\authorrunning{Miu Lun Lau et al.} 
%
\tocauthor{Miu Lun Lau, Min Long, Jeff Terry}
\institute{Boise State University, Boise ID USA,\\
\email{andylau@u.boisestate.edu, minlong@boisestate.edu}
\and
Illinois Institute of Technology, Chicago, Illinois,\\
\email{terryj@iit.edu}\\
}

\maketitle              

\begin{abstract}
We introduce a Genetic Algorithm (GA) based, open-source project to solve multi-objective optimization problems of materials characterization data analysis including EXAFS, XPS and nanoindentation. The modular design and multiple crossover and mutation options make the software extensible for additional materials characterization applications too. This automation of the analysis is crucial in the era when instrumentation acquires data orders of magnitude more rapidly than it can be analyzed by hand. Our results demonstrated good fitness scores with minimal human intervention.
\keywords{Genetic Algorithm, EXAFS, X-Ray}
\end{abstract}

\section{Introduction}
\par Advanced data analysis techniques for materials characterization data sets are of growing interest to the material science community due to its capability in detecting and potentially predicting the properties of materials.
The performance of this process is limited by two main factors: the significant inputs from users needed to retrieve important structural parameters and the quantity of high quality data that is now collected due to advances in instrumentation. The experiential data collected from modern instruments are orders of magnitude larger than can be analyzed by trained personnel. For example, the Fourth Generation Synchrotron Light Sources that are coming online are expected to produce data at rates 2-3 orders of magnitude greater than the current collection rates that are as high as 6 GB/s \cite{Blaiszik2019}. Advances in processing are necessary to make in-situ real-time characterization of materials feasible. Processing large quantities of data can result in reproducibility problems \cite{major2020} when inexperience users make errors that then slowing research productivity, which discourages the quest for research excellence, and inhibits effective technology transfer and manufacturing innovation.   

In order to address the need to analyze massive datasets both quickly and accurately through solving multi-objective optimization problems, we have been developing an open-source code base featuring a Genetic Algorithm (GA) that can analyze a variety of materials characterization data types with minimal human input to retrieve important parameters \cite{Terry2021}. Although it is still under active development, it has already demonstrated the ability to automatically analyze extended X-ray Absorption Fine Structure (EXAFS) data, giving a reproducible description of the local atomic structure of materials.

The GA-based EXAFS analysis package called \texttt{EXAFS Neo} \cite{Neo} is written primarily in Python. It requires the installation of \texttt{Larch} software package \cite{newville2013larch} and utilizes a version of \texttt{FEFF8.5l} \cite{Rehr2010} code contained within \texttt{Larch} for calculating the initial scattering paths. The code was not parallelized and we found that parallelization was really not needed for effective use as an analysis tool. However, we strongly recommended that users execute multiple calculations of the same data set simultaneously for further error analysis. The included graphical user interface (GUI) allows for populating multiple parameter sets for simultaneous exploration of parameter space. Several tutorial video demonstrations of the \texttt{EXAFS Neo} package are available for viewing at the package download website \cite{Neo}. Due to the modular design, this open-source software can be extended to other materials characterization datasets. Our recent efforts include extensions for the analysis of X-ray Photoelectron Spectroscopy (XPS) and nanoindentation data. 
 
We will briefly introduce the materials characterization analysis in Section II, with a focus on the group of parameters not commonly seen in other GA codes. The design and implementation of GA for this problem is given in Section III. We evaluate the experimental results and give performance analysis of the code in Section IV. Finally, a summary of the work is given in Section V.

\section{Analysis of Materials Characterization Data}

\par We use EXAFS as an example to demonstrate the process of materials characterization analysis. The principles discussed here are directly applicable to the anslysis of other characterization tools such as XPS and nanoindentation. EXAFS is used to study local structure around a specific element on the atomic scale. It can be applied to a wide range of materials including liquids, amorphous solids, and crystalline materials. The absorption spectrum, $\mu(E)$, with energy, $E$, of X-ray photons can be represented as a combination of the background $\mu_{0}$ absorption and oscillations due to scattering events $\chi(E)$, i.e., $\mu(E) = \mu_{0}(1+\chi(E))$. This equation is usually rewritten using the photoelectron wavenumber $k = [{\frac{2m}{\hbar^{2}}(E-E_{0})}]^{1/2}$, where $m$ is the electron mass, $\hbar$ is the Plank's constant, and $E_{0}$ is called the energy of the absorption edge. This results in the following equation for $\chi(k)$: 
\begin{equation}
\chi(k) = \frac{\mu(k)-\mu_{0}(k)}{\mu_{0}(k)}.
\end{equation}

\par The oscillations $\chi(k)$ result from the interference of  scattered photoelectron waves. Scattering can involve \textit{single} or \textit{multiple} scattering events. In \textit{single} scattering events, the electron wave scatters from only one neighboring atom before returning to the source atom. In \textit{multiple} scattering events, the wave may travel to multiple nearby atoms through a variety of paths before returning to the source atom. The trajectories of the photoelectron waves are described as \textit{paths}. These parameters describing these paths become the gene components of the chromosomes in the GA code.

\par The measured X-ray absorption spectrum (XAS) \cite{Bunker2009} consists of two distinct regions: the X-ray Absorption Near Edge Structure (XANES) region
and the Extended X-ray Absorption Fine Structure (EXAFS) region, which begins at the end of the XANES region and extends beyond the absorption edge until the oscillations damp out. The high energy endpoint is dependent upon the nature of the scattering atoms. The EXAFS region is used to study interatomic distances, coordination numbers, and lattice dynamics, from which one can often infer the surface chemistry of complex systems.

The interference that leads to the measured EXAFS oscillations is due to the interactions of the emitted photoelectron wave, ejected from inner core shells by resonant radiation, with the waves scattered by neighboring atoms. These scattered electron waves modulate the wavefunction of the original photoelection. If the interference is destructive, a photon cannot be absorbed because the electron wave cannot be created. Conversely, if the interference is constructive, more photons are absorbed. 
Sayers et al. \cite{sayers1971new} was first to invert the measured experimental EXAFS data into quasi-radial distribution functions using a simple point scattering theory. Extensions to this simple methodology result in the full EXAFS equation:

\begin{equation}
\chi(k) = \sum_{i (paths)} \frac{(S_0^2 N_i) F_i(k)}{k R_i^2} e^{-2\sigma_i^2k^2}e^{-2R_{i}/\lambda(k)} \sin[2kR_{i} + \phi_i(k)+\delta_c(k)],
\label{eq:EXAFS_EQ}
\end{equation}

\noindent
where $i$ represents an individual scattering path, $R_{i}$ is half of the scattering distance, $\phi_{i}(k)$ is phase shift due to scattering, and $\delta_{c}(k)$ is the phase shift due to the potential of the absorbing atom. $N_{i}$ is the degeneracy of the path, and $S_{0}^{2}$ is the amplitude reduction factor that arises from quantum mechanical considerations. $S_{0}^{2}$ and $N_{i}$ are coupled together to describe the amplitude of each scattering path. $F_{i}(k)$ is the effective scattering amplitude for the waves in path and $\sigma^{2}$ is the Debye-Waller factor, which accounts for the thermal and static disorder. 

We can observe from Eqn. (\ref{eq:EXAFS_EQ}) that the EXAFS signal is a summation of sinusoidal waves of varying amplitudes from scattering paths containing the neighboring atoms. The waves are inherently spherical in nature and are affected by the type of atoms, temperature, neighboring atoms, as well as inelastic losses. The scattering events $\chi(k)$ typically decay rapidly with increased wave number $k$, which leads to lower signal to noise values at high-k. The $\chi(k)$ function is often weighted by $k^{2}$ or $k^{3}$ to highlight the contribution from the high-k regions of the spectrum. Analysis of the spectrum of $k^2\chi(k)$ in terms of $k$ is known as K-space analysis.  

\begin{figure}[htbp]
	\centering
	\includegraphics[trim=0 0 0 0, clip,width=0.9\linewidth]{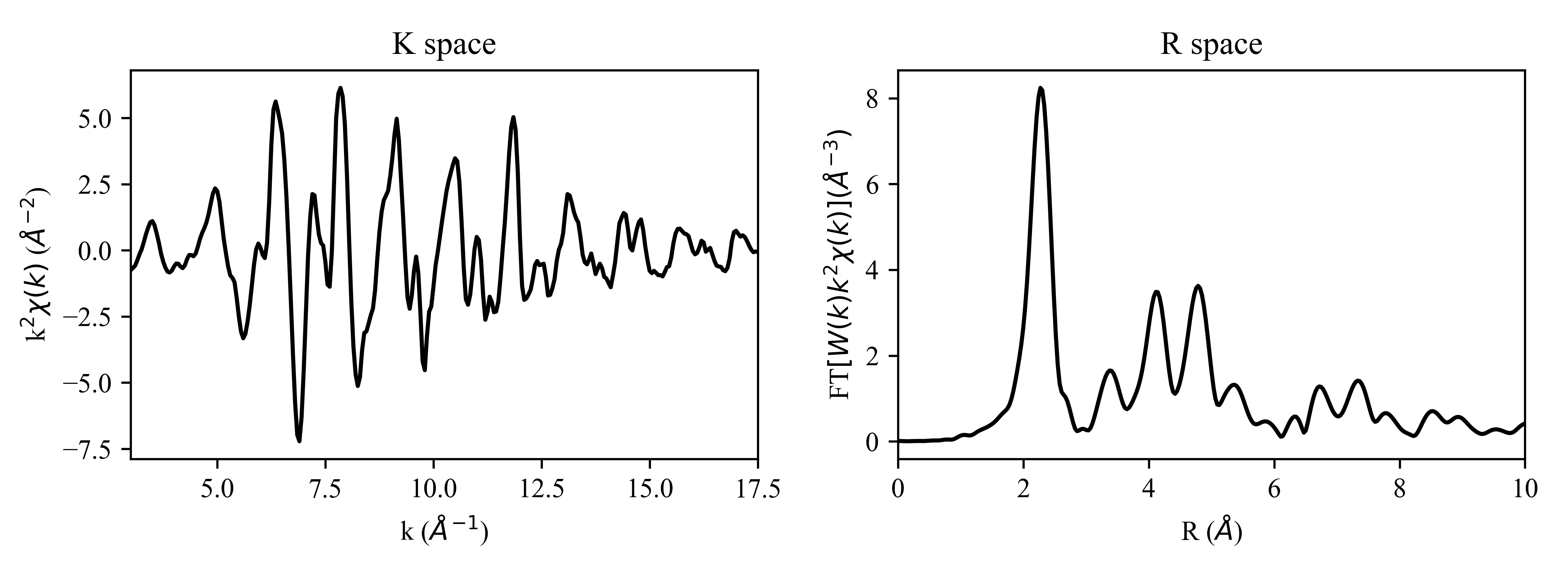}
	\caption{EXAFS spectra of copper metal of K edge in K and R-space. The data was taken from XASLAB \texttt{https://xaslib.xrayabsorption.org/spectrum/91/}.}
	\label{fig:exafs_spectra_comb}
\end{figure}

A Fourier Transformation can be applied to the K-space data which leads to R-space EXAFS data. This R-space data is not a true radial distribution function, unfortunately. It can be used to limit the analysis to a specific range of individual scattering paths. An example of K and R-space EXAFS data is shown in Figure \ref{fig:exafs_spectra_comb}. The transformed spectra is:
\vspace{-0.1in}
\begin{equation}
\chi(r) = \frac{i\delta k}{\sqrt{\pi N_{max}}}\sum_{n=1}^{N_{max}}\chi(k)\Omega(k)k_{n}^{w}\exp(2i\pi n/N_{max})
\end{equation}

\noindent where $\delta k$ is the spacing in the k-space spectrum; $N_{max}$ is the array size; $\Omega(k)$ is the input window function, used to reduced oscillation from truncation of the input spectrum; and $k_{n}^{w}$ is the k weight applied to the data.

\section{Design and Implementation of GA Analysis Code}

\par Conventional analysis of EXAFS can be very difficult depending on the number of scattering paths needed to fit the experimental data. Each path is characterized by a set of four parameters:  $[(S_0^2 N_i)$, $\Delta E_0$, $\sigma^2_{i}$, $R_{i}]$. We use $\Delta E_{0}$ to correct energy mismatch of the absorption edge between experiment and theory. It should be identical for all scattering paths from the same calculation. Many software packages have been developed for EXAFS analysis, including \texttt{Larch} \cite{newville2013larch}, \texttt{Demeter} \cite{Ravel2005}, \texttt{WinXAS} \cite{WinXAS}. However these tools require significant knowledge of condensed matter physics, and can be difficult for novice users to use. They can also be quite time-consuming to use.

\par To address some of these issues, we have been developing an automated Materials Characterization Software package for EXAFS, XPS, and nanoindentation analysis. The code is GA-based for such a multi-objective optimization problem. We should point out we are not the first to apply GA to EXAFS analysis \cite{bunker2005new}. However a comprehensive study (e.g., crossover and mutation options) of GA algorithms and their effects on uncovering the parameters for materials characterization analysis have not been studied. This is the goal of this paper.

\par GA is a heuristic optimization method inspired by the Darwinian theory of evolution \cite{holland1992genetic}\cite{goldberg1988genetic}. At the start of the algorithm, a population consisting of $n$ temporary solutions (individuals) is generated randomly throughout the solution domains. Each solution is considered as a chromosome consisting of the parameters of interest and each parameter represents a gene of the chromosome. The GA evaluates the fitness of each solution in a population using a fitness function to determine the evolution of the next generation of solutions. To improve the accuracy of the final solution and the convergence rate of the involved iterations, a number of evolutionary inspired operators (e.g., crossover, mutation) are applied to each solution throughout subsequent generations. Since GA is also a stochastic process, it requires \textbf{multiple} runs of identical parameters to evaluate if the resulting solution reaches a global optimum. Compared to other search methods, GA operates by performing a multi-dimensional search, and encourages information exchange among the different solutions. We will briefly describe how various operators affect the optimization process and our implementation for EXAFS analysis. The literature of \cite{eiben2003introduction} is listed for reference on general principles of GA.

\subsection{Chromosome Representations of $[ S_{0}^{2}, \Delta E_{0}, \sigma^{2}, \Delta R_{i} ]$}
\par The conventional GA cannot be applied directly but has to be customized for automated materials characterization analysis due to the reasons discussed in Section II.  Unlike the conventional GA, where binary strings are used as individual chromosome in each individual, our software uses a floating point representation of a set of parameters as individual components: $[ S_{0}^{2}, \Delta E_{0}, \sigma^{2}, \Delta R_{i} ]$. $\Delta R_{i}$ is used instead of $R_{i}$ because we must find how to modify the theoretical paths to reproduce the experimental data. $S_{0}^{2}$ other than ($S_{0}^{2}N_{i}$) is adopted because we assume the degeneracy parameter $N_{i}$ is static and taken from the ideal atomic structure of the compounds under evaluation. In addition, all our parameters are constrained to valid physical ranges to prevent unreasonable values from being introduced, such as non-positive amplitude or Debye-Waller factors. When the number of ``paths" of interest are determined, e.g., $n_{path}$, each individual in evolutionary population will consists of \textbf{multiple} (e.g.,$n_{path}$) sets of parameters.

\subsection{Crossover and Mutation}

\par To allow convergence toward the global optimal solution, the best solutions of each generation are selected to be parents of the future populations, mirroring observation seen in nature. The selection process is based upon the fitness values in the population. We employed a selection approach similar to rank selection, where each solution was sorted based on their fitness values, from best performing to the least. A certain percentage of the best performing solutions are selected for retention in the populations. To ensure biodiversity and reduce the risk of getting trapped in local extrema, a percentage of random solutions is generated as well. The main challenge regarding the march toward the optimal solution involves maintaining a balance between population diversity and selective pressure. A large population diversity leads to slow convergence on the optimum solution, while a low population diversity leads to premature convergence. 

\par \textit{Crossover} or \textit{Recombination} is described as the operation of combining parental materials of two or more solutions during which the information is inherited and exchanged to produce a new solution. In nature most species have two parents but in GA the crossover operations can extend to more than two parents. Following the crossover operator, two individuals are selected to generate subsequent individuals throughout the GA. There are numerous techniques for the crossover operator in the literature.\cite{eiben2003introduction} For example, in the single-point cross over, the chromosomes of two parent solutions are swapped before and after a single point. In the double-point crossover, however, there are two cross over points and the chromosomes between the points are swapped only. We developed three crossover methods in our GA code and tested them for EXAFS analysis. The details of results will be discussed in Section IV. 

\par \textit{Mutation} operators modify existing solution by disturbing them by random chance. The mutation usually occurs in the ``gene" level. For example, an offspring in binary representation of ``0101" can become ``0111" depending upon the probability of mutation or mutation rate. The exploration of additional portions of phase space by the mutation operator can help to find the global optimal solution by allowing the escape from local optimal solutions. The mutation rate is usually set to be very low, but allows for perturbations in the ranges of values. Conventionally, the mutation operator remains fixed throughout the entire optimization process. This can lead to ill-condition where the mutation operator is not sufficient to steer the solution out of a local extreme, which leads to premature convergence of the solution set. A self-adaptive mutation operator can mitigate premature convergence by increasing or decreasing the chance of mutation at each generation, which allows the algorithm to continuously refine the search area.

\par In our GA code, we implemented an algorithm based on the Rechenberg 1/5 success rule \cite{rechenberg1994evolutionsstrategie}, which increases the mutation probability ($\sigma$) based on the ``success ratio," $S_{i}$, at the current generation. It is defined as the probability of generating an improved fitness value compared to the previous population. This probability can be further extended to the \textit{Crossover} operator as well but was limited to the \textit{Mutation} operator in this work. The mutation rate will increase in subsequent generation if $S_{i}$ is greater than 1/5, and decrease if it is less.

\subsection{Fitness Calculation and Exit Conditions}
\par The fitness calculation is used to compare the quality of individuals within the given population. Typically, the fitness is computed every generation to ensure the solution is converging toward the optimal. However, the calculation can be very expensive and most approaches aim to minimize the number of fitness function calls. The fitness function employed for EXAFS is determined by computing the differences between the GA model and experimental data at each data point using a $\chi^{2}$ model:
\begin{equation}
\label{eq:chiR2}
\chi^{2} = \frac{N_{indep}}{N}\sum_{i=1}^{N}\frac{\left(y^{\rm{model}}_{i}-y^{\rm{data}}_{i}\right)^{2}}{\epsilon_{i}^{2}}.
\end{equation}

\noindent $y_{i}^{\rm{model}}$ represents the model data (e.g., $\chi(k)$) at $i$, $y_{i}^{\rm{data}}$ represents the experimental data, $N_{indep}$ is the number of independent points, $N$ is the total number of data points, and $\epsilon_{i}$ is the measure of uncertainty at each data point. Depending upon the collection methodology of the EXAFS data, the ratio $N_{indep}/N$ can range from $1/10$ when the EXAFS signal is over-sampled to speed data collection to 1 when the data is collected stepwise. In fact, this ratio can vary across the energy range of the collected EXAFS data. 
For other materials characterization analysis techniques like XPS and nanoindentation, we can utilize similar fitness functions (e.g., Eqn. \ref{eq:chiR2}) but with the appropriate theoretical functions used to replace Eqn. \ref{eq:EXAFS_EQ} to construct the GA model. This modularized approach makes our GA-based code an excellent platform for expansion to other applications.

\par The principles of GA present the reasoning that an optimal solution can be reached after enough iterations. However, practical applications may deviate from this ideal case because it is possible that the optimal solution has not been found before iterations stop. Therefore, it is important to implement proper exit conditions to terminate the running of the code. Our code uses two method for determining when the algorithm reaches exit conditions. The first one is to set a maximum number of iteration by users. The second method is more adaptive. When the solution doesn't improves for a number of generations, we exit with the assumption that an optimal has been reached. In our software, we provided both methods of controlling the exit condition of the code.

\subsection{Error Analysis}

\par GA is a probabilistic optimization method where the solution can not be expected to be identical over repeated runs. It is \textbf{critical} to perform a set of individual, independent runs (e.g., 50) with varying conditions (e.g., population size, generation number, mutation rate) to evaluate the accuracy of the solution and determine the range of errors. In our GA code, we use the Global Random Analysis method \cite{redhouse2017uncertainty} to estimate errors of the GA-optimized parameters. The error of each individual parameter (e.g., $S_{0}^{2}, \Delta E_{0}, \sigma^{2}, \Delta R_{i}$) in the simplest of X-ray absorption cases can be quantified using a Poisson distribution and the standard deviation is roughly equal to the square root of the absorption duration time \cite{meyer1965introductory}. It is worth noting that this quantification can only be used when the sample is uniform in composition and thickness, and cannot be applied to heterogeneous samples or samples which have suffered any irradiation damage. However, these limitation do potentially affect the error analysis of this methodology as they do in conventional EXAFS analysis.

During each generation, the individuals of the best fitness value are stored and used to construct the covariance matrix which contains the error for each parameter in the solution. Random perturbations of three parameters are selected in a bounded range: population size (100-5000), total number of generations (10-50), and mutation rate (0\%-100\%). For example, one individual run of 50 runs could randomly select a condition in the bounded ranges like population size 200, generation number 30, mutation rate 20\% and other runs may select other conditions. The GA code is run using each set of new starting conditions. The combined results are utilized for global random analysis.

\section{Experiment Results}

\par The majority of the code is implemented and written in Python. All our experiments were conducted on the Idaho National Laboratory High Performance Computing (HPC) cluster `Sawtooth' which is comprised of Xeon Platinum 8268 Processors. Each optimization was  repeated 100 times to compute the standard deviations and error matrix, and for each experiment we requested four CPU cores and 12.0 GB of memory. To validate our GA code before applying it to analysis of real experimental EXAFS data, we first applied the code to a set of synthetic data, which is generated from five known scattering paths. Gaussian noise with a signal to noise ratio of 20 was added to the synthetic spectra. 
\begin{table*}[t]
	\centering
	\footnotesize
	\caption{GA fitted parameters and errors from the synthetic data set.}
	\begin{tabular}{cccccccccc}
		\toprule
		Path & $N$ & \multicolumn{2}{c}{$S_0^2$} & \multicolumn{2}{c}{$\Delta$E$_0$ (eV)} & \multicolumn{2}{c}{$\sigma^{2}$ (Å$^2$)} & \multicolumn{2}{c}{$\Delta$R (Å)} \\
		\cmidrule(lr){3-4}\cmidrule(lr){5-6}\cmidrule(lr){7-8}\cmidrule(lr){9-10}
		\#   &     & Model  & True                & Model          & True                   & Model  & True                & Model         & True   \\
		\midrule
		1 & 12\ & 0.62$\pm$0.03 & {0.62} & -0.3$\pm$0.7 &{-0.91} & 0.0041$\pm$0.0004 & {0.004} & 0.0512$\pm$0.003 & {0.05} \\
		2 & 6\ & 0.72$\pm$0.08  & {0.66} & -0.3$\pm$0.7 &{-0.91} & 0.0015$\pm$0.0007 & {0.001} & 0.0118$\pm$0.005 & {0.01} \\
		3 & 48\ & 0.5$\pm$0.2   & {0.74} & -0.3$\pm$0.7 &{-0.91} & 0.0102$\pm$0.0044 & {0.014} & 0.0615$\pm$0.032 & {0.08}  \\
		4 & 48\ & 0.3$\pm$0.2   & {0.45} & -0.3$\pm$0.7 &{-0.91} & 0.0068$\pm$0.0043 & {0.009} & 0.0467$\pm$0.043 & {0.00} \\
		5 & 24\ & 0.22$\pm$0.08 & {0.14} & -0.3$\pm$0.7 &{-0.91} & 0.0081$\pm$0.0024 & {0.005} & 0.0538$\pm$0.012 & {0.05} \\
		\bottomrule
	\end{tabular}
	\label{tb:Synthetic Data}
\end{table*}

Tab. \ref{tb:Synthetic Data} shows a set of GA fitted parameters with the corresponding errors generated from the synthetic data. The true values are also listed for comparison. Since the scattering with single path has the shortest traveling distance, it has the largest contribution to the absorption spectrum and is of particular interest. It can be seen from the first row (Path 1), all of the fitted parameters: $S_{0}^{2}$, $E_{0}$, $\sigma^{2}$, and $\delta_{R}$ match extremely well to the true values, indicating a high confidence of accuracy using our GA model. For the weaker scattering paths, the errors are found to be larger but are comparable to conventional analysis methods.

\begin{figure}[h]
	\centering
	\includegraphics[width=0.9\linewidth]{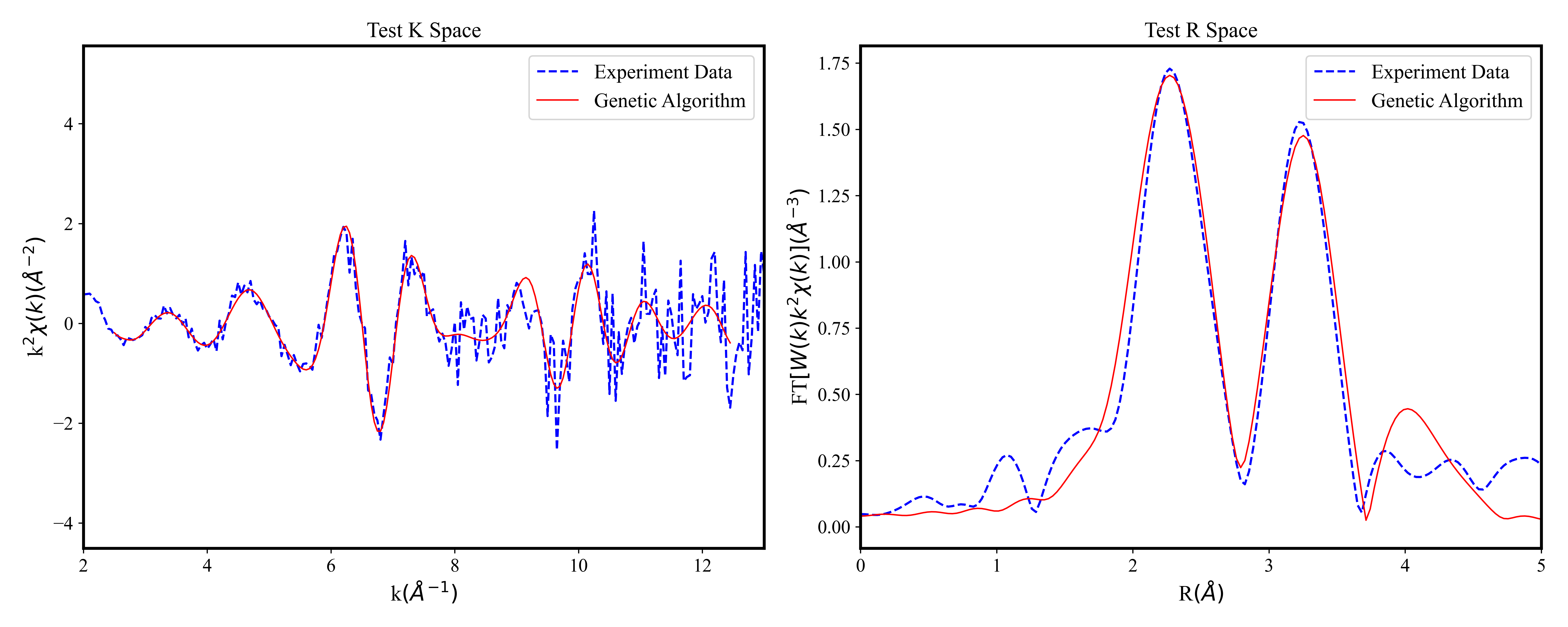}
	\caption{EXAFS spectra fits of the synthetic experimental spectra with added Gaussian noise. Left: Spectra were fitted in K-space over the region from 2.5 to 12.5 $\AA^{-1}$. Right: Spectra in R space in a range from 0 to 5 $\AA$.}
	\label{fig:exafs_spectra}
\end{figure}

Fig. \ref{fig:exafs_spectra} shows the GA fitted spectra in both K-space and R-space in comparison with the true synthetic data. The included Gaussian noise was higher than what is usually observed in real experimental EXAFS data. This choice was made to test the performance of our GA code under extreme conditions. It can be seen that the GA code was able to obtain good matches to the actual values, especially in lower $k$ ranger, which contribute to the fitted parameters with the highest level of signal to noise ratio.

\begin{figure}[htbp]
	\centering
	\includegraphics[width=3in]{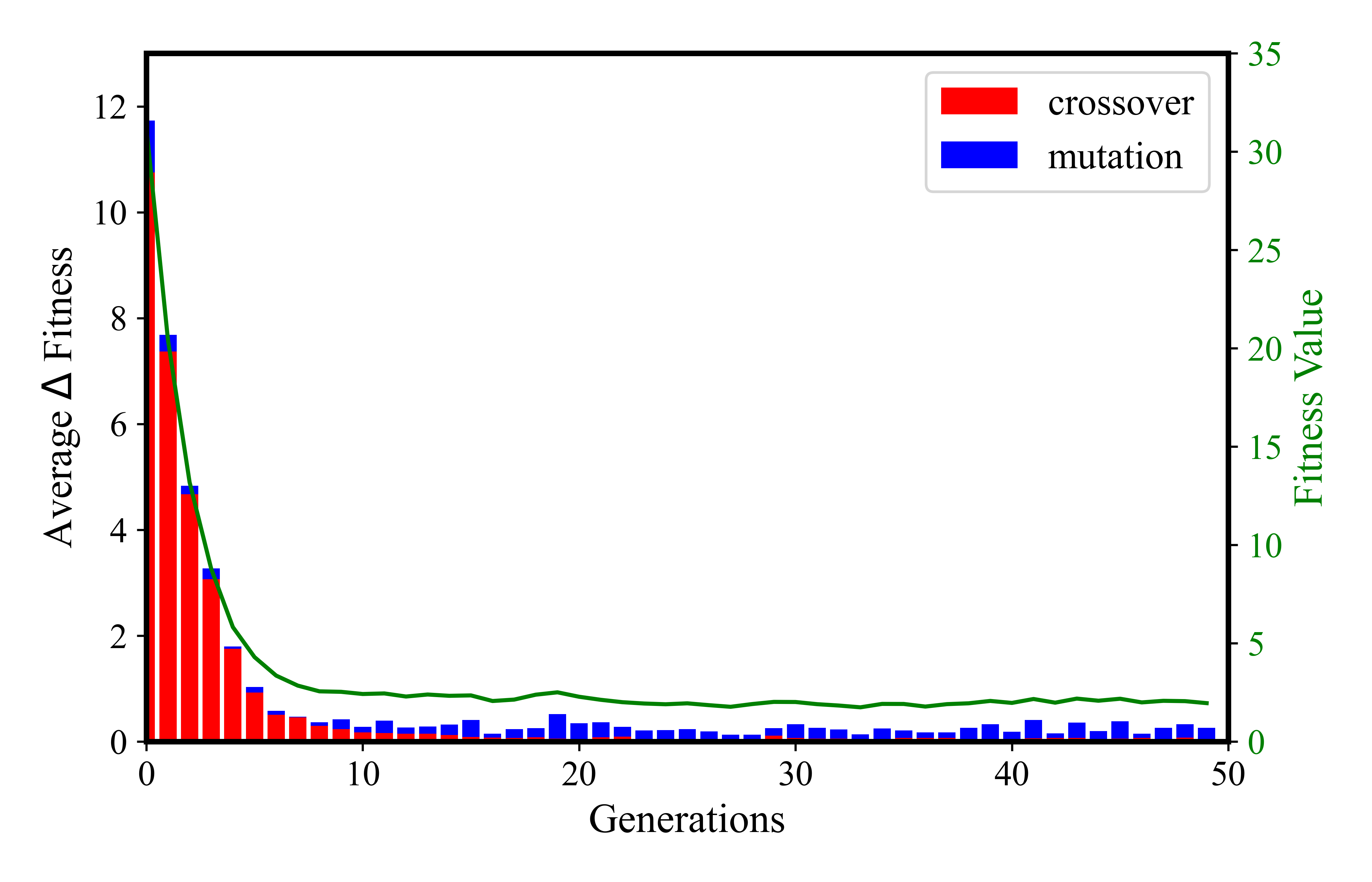}
	\caption{The effect of the crossover (red) and mutation (blue) operators on the overall calculation of fitness value.}
	\label{fig:D_fitness}
\end{figure}

\par Fig. \ref{fig:D_fitness} shows the effects of varying both the mutation and crossover operators, to explore their contributions to the change of the fitness value. We can observe the average fitness value significantly decreases (i.e., betters) along with the evolution of generations. The crossover operator has the largest contribution to the improvement of the fitness value in the early stages of the evolving generations, which can be seen from the significant heights of red bars in Fig. \ref{fig:D_fitness}. The mutation operator contributes less in general compared to the crossover operator in early stages and even has zero contribution in a few cases. However, the mutation operator contribution dominates at the late stages of the evolution. This is because the mutation probability increases as the generations evolve under the Rechenberg algorithm, which leads to an increase in amount of phase space probed by the mutation that occur. Mutation is also used to check if the solution is in global optimal.  

\par Next, we will employ three metrics to evaluate the quality of our fits in both K and R spaces: Coefficient of Variation (R2), Mean Absolute Error (MAE), and Root Mean Square Error (RMSE) \cite{bevington1993data}. R2 measures the residuals of spectra, MAE measures errors within the same unit spectrum but gives similar weights to outliers, and RSME is used to address amplitude errors from outliers. 

\subsection{Crossover Analysis and Mutation Analysis}
\par We provided three crossover methods in our GA code. Users can examine their effects and select the best option for their specific problems. The first method ``uniform random crossover" utilizes uniform crossover where each gene is selected randomly from either parent with equal probability. The second method ``AND crossover" utilizes a mixing rule where each gene in child is a result of an AND logic operation on to each gene from its parents. The third method``OR crossover" utilizes the similar mixing rule using an OR logic operator. 

\vspace{-0.1in}
\begin{table}[htbp]
	\centering
	\caption{Error analysis for three crossover and three mutations methods.}
	\begin{tabular}{ccccccc}
		\toprule
		       & \multicolumn{3}{c}{Crossover} & \multicolumn{3}{c}{Mutation} \\
		\cmidrule(lr){2-4}\cmidrule(lr){5-7}
		Method& 1(uniform random)  & 2(AND) & 3(OR) & 1(maximum) & 2(nested) & 3(metropolis) \\
		\toprule
		R2-K	& 0.99856 & 0.95530 & 0.94321 & 0.99900 & 0.98625 & 0.99920 \\
		R2-R	& 0.99661 & 0.98541 & 0.97431 & 0.99632 & 0.99163 & 0.99656 \\
		\midrule
		MAE-K	& 0.03912 & 0.17298 & 0.03345 & 0.03389 & 0.11349 & 0.03047 \\
		MAE-R	& 0.03455 & 0.04561 & 0.04733 & 0.03700 & 0.04119 & 0.03611 \\
		\midrule
		RMSE-K	& 0.00231 & 0.05333 & 0.00255 & 0.00175 & 0.05250 & 0.00142 \\
		RMSE-R	& 0.00211 & 0.01044 & 0.00355 & 0.00264 & 0.00604 & 0.00247 \\
		\bottomrule
	\end{tabular}
	\label{tb:crossover_mutation}
\end{table}

\par Tab. \ref{tb:crossover_mutation} shows that the first method returns the highest contribution to the true value in both K and R spaces. The second and third methods may produce lower accuracy than the uniform random crossover in our application. This is because our solution set only contains a limited number of four parameters, therefore it is ideal to use uniform random crossover to promote information exchange and diversity as much as possible between individuals. For this reason, we adopted the uniform random crossover in the following experiments.  

\par We also provided three mutation methods in our GA code for users. They all use the same starting mutation probability at each generation $\sigma_{i}$ and compare it with some random number to determine if a new mutation is necessary. The first method ``maximum mutation" generates a complete new individual in units of the parameters if a random generated number reaches the mutation probability. It is a simple method and maximizes the number of possible mutations in the population. It ensures that sufficient genetic diversity is introduced within the overall population.  The second method ``nested mutation" introduces a secondary random number to control the actual mutation. The new individual can only be generated in units of gene when the both random numbers are less than the mutation rate at each generation. This method fits a more traditional GA where the actual mutation rate for each gene is typical set very low (e.g., 1 to 5\%).   

\begin{minipage}[t]{.46\textwidth}
	\vspace{-0.3in}  
	\begin{algorithm}[H]
		\centering
		\caption{Mutation method 2: Nested Mutation. Note: Removing lines 4,5,6,8,9 can lead to Mutation method 1: Maximum Mutation.}
		\begin{algorithmic}[1]
			\renewcommand{\algorithmicrequire}{\textbf{Input:}}
			\renewcommand{\algorithmicensure}{\textbf{Output:}}
			\REQUIRE Mutation Rate $\sigma$, Individual $P_{i}$
			\ENSURE  Individual $P_{i}$
			\FOR {$P_{i}$ in Populations ($P$)}
			\STATE Generate a number $x$ in [0..100] 
			\IF{$x <\sigma$}
			\FOR {number of variables in each scattering paths}
			\STATE Generate a number $y$ in [0..100]
			\IF{$y < \sigma$}
			\STATE Generate new individual
			\ENDIF			
			\ENDFOR
			\ENDIF
			\ENDFOR
		\end{algorithmic}
		\label{alg:mut_1}
	\end{algorithm}
\end{minipage}%
\hfill
\begin{minipage}[t]{.46\textwidth}
	\vspace{-0.3in}  
    \begin{algorithm}[H]
	\caption{Mutation method 3: Metropolis Mutation}
	\begin{algorithmic}[1]
		\renewcommand{\algorithmicrequire}{\textbf{Input:}}
		\renewcommand{\algorithmicensure}{\textbf{Output:}}
		\REQUIRE Mutation Rate $\sigma$, Individual $P_{i}$, Fitness $f_{i}^{orig}$
		\ENSURE  Individual $P_{i}$
		\FOR {$P_{i}$ in Populations ($P$)}
		\STATE Generate a random $x$ in [0..100] 
		\IF{$x< \sigma$}
		\STATE Mutate the Scattering Paths
		\STATE Calculate New Fitness $f_{i}^{mut}$
		\IF{$f_{i}^{mut}< f_{i}^{orig}$ }
		\STATE Accept the Mutation
		\ELSIF {$\exp(-(f_{\mathrm{mut}} - f_{\mathrm{orig}})/K(i))< t$}
		\STATE Accept the Mutation
		\ELSE
		\STATE Reject the Mutation
		\ENDIF
		\ENDIF
		\ENDFOR
	\end{algorithmic}
	\label{alg:mut_2}
    \end{algorithm}
\end{minipage}

The third method ``metropolis mutation" is modeled after the Metropolis-Hasting algorithm \cite{chib1995understanding}, which rejects the mutation if the fitness value is less than current fitness value, or accepts it if the following criterion is satisfied:
$\exp(-(f_{i}^{\mathrm{mut}} - f_{i}^{\mathrm{orig}})/K(i))< t$.
The $f_{i}^{\mathrm{mut}}$ and $f_{i}^{\mathrm{orig}}$ are fitness values after and before the mutation, $t$ is a random uniform value $\in [0,1]$, and where $K_{i}K(i) = - \delta_{f}/\ln(1-i/i_{max})$ is a parameter called the cooling rate. The $\delta_{f}$ is the absolute difference in fitness value between subsequent generations, $i$ is the current generation number, and $i_{max}$ is the maximum generation. This cooling rate allows the possibility of accepting parameters outside of the predefined range, while assuming that the mutation rate will decrease at a linear rate.

Table \ref{tb:crossover_mutation} compares the different mutation methods. We can see that the value of R2 for mutation method 1 and method 3 is very similar, but the fitness value for the method 2 performed the worst.  This indicates that analyzing a relatively small parameter set in the EXAFS application would require a large mutation rate to enhance the mutation possibilities and ensure the diversity of the population.

\subsection{Cut off Analysis for Selecting Scattering Paths}

\par One of the main difficulties in obtaining an accurate fit using GA for EXAFS analysis lies in determining the number of scattering paths that may potentially be observed in an EXAFS measurement and selecting from that set the actual paths required to replicate the experimental results. The list of most significant paths can be difficult to obtain due to the effectively infinite number of potential path combinations. On the other hand, there is no universally optimized fixed set of paths for all applications and it is important to deselect insignificant scattering paths with low contribution to the spectra. 
The analysis tools must be able to handle these cases. 

\par We have developed a process to analyze the contribution from potential scattering paths and uncover the scattering paths with the largest contribution to the spectra. We first calculate the integrated area below the spectrum curve and the contribution from each individual path. We selected the paths that contribute more than a user-selectable, pre-defined cut off percentage of these areas (e.g., 1\%). In this manner, we obtain a list of significant paths for further analysis. 

\par Fig. \ref{fig:cutoff} shows $\chi^{2}$, from Eq.~\ref{eq:chiR2}, as a function of the cut off percentage.  We analyzed four data sets of copper K-edge EXAFS spectra from metallic copper collected at temperatures of 10K, 50K, 150K and 298K to test the path selection method based on the cut off area percentage. The experimental data were obtained from XASLAB. The initial number of scattering paths that we employed was 42 which represented a full set of scattering paths with distances from 2.5527 \AA\ to 7.6580 \AA. These paths were used to fit the experimental data over the K-space range from 3 $\AA^{-1}$ to 17.0 $\AA^{-1}$.  After performing the cut off calculation, a subsequent optimized set of paths was obtained and applied to a second round of GA optimization that only used the optimized scattering path list and excluded all paths with insignificant contributions to the measured spectrum. 

\par The results of utilizing the cut off can be evaluated by using the final fitness value score (i.e., $\chi^{2}$). However, we must strike a balance between the number of scattering paths and the accuracy of the final fitness value. We tested our algorithm by using multiple cut off percentages to observe the effects on the final average fitness value. We selected seven different cut off percentages: 10\%, 5\%, 2\%, 1\%, 0.67\%, 0.5\% and 0.3\%. 

We observed that when the cut off percentage decreased from 10\% to 1\%, the average $\chi^{2}$ score decreased (indicating improvement) for all of temperatures.  When the cut off percentage went below 1\%, the $\chi^{2}$ score tended to increase due to the increased number of paths that were included in the calculation. The increased number of paths led to insignificant changes in the overall fitness score. In our experiments, we found that the best cut-off ratio was in the range of 1\%, although we allow users to select other cut off values for their applications. 

\subsection{Computational Performance}

\par It is worth noting that there is no emerging need to parallelize our GA code since the same data set must be run/analyzed multiple times (e.g., 100) to gauge the errors compared to the experiments. However, parallelization is applicable if the number of IO operations is reduced significantly. In our current GA code, evolutionary operators perform very frequent IO operations. 

\par We measured the scalability of our GA code in terms of the number of scattering paths. There are two major factors which can affect the complexity of the computation.  The first is the number of sample points in the spectra, which is usually determined by the experimental conditions of the measurement, or instrumentation setting. The second is the number of scattering paths selected by users to represent the exploration range of atomic structure of interests. Fig. \ref{fig:TPG} shows the average time spent per generation as a function of the number of selected paths. We can see that the algorithms scales very well, with a time complexity $\mathcal{O}(n)$. 

\begin{minipage}[t]{.46\linewidth}
\vspace{-0.3in}
\begin{figure}[H]
	\centering
	\includegraphics[width=\linewidth]{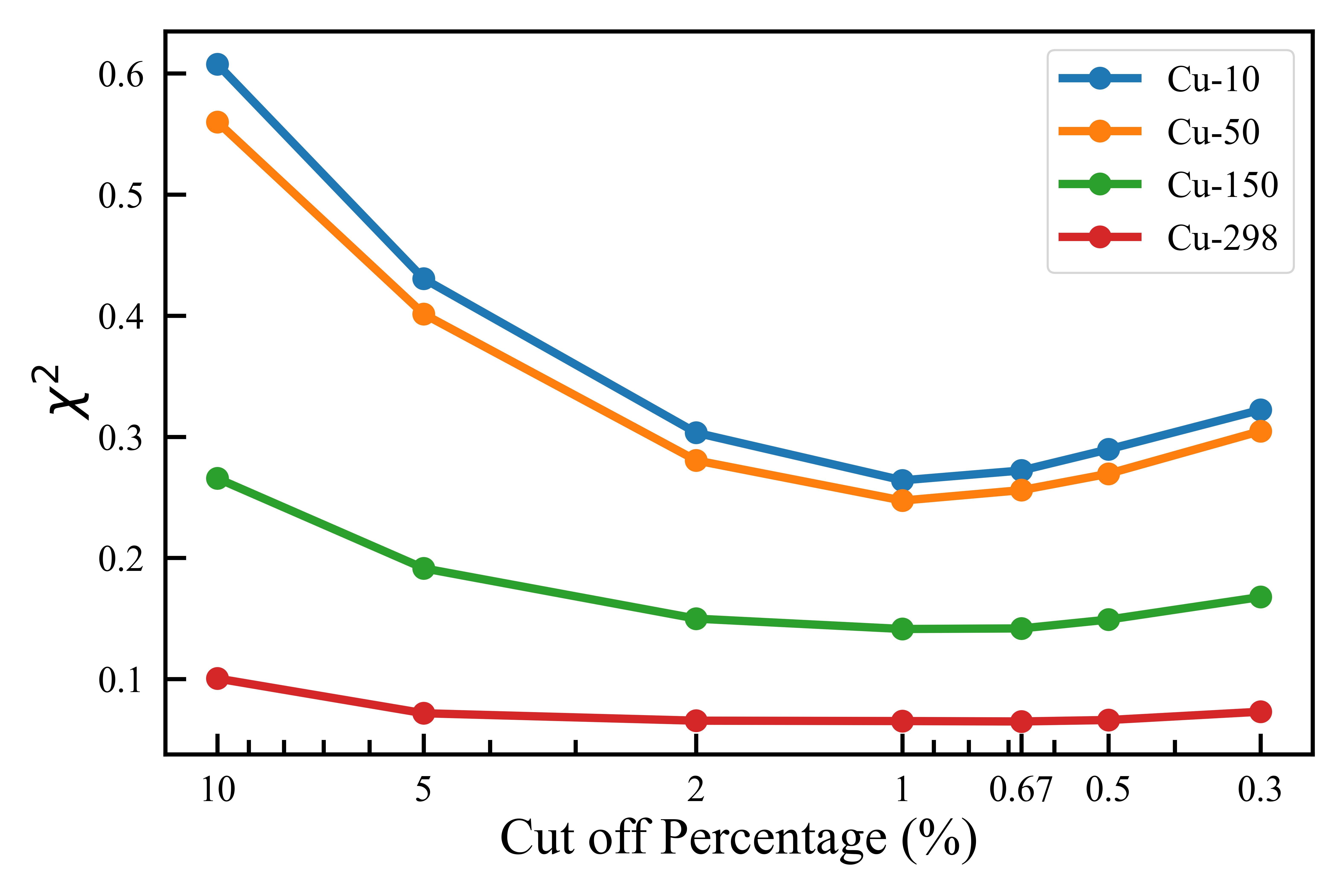}
	\caption{$\chi^{2}$ as a function of the cut off percentage for Cu foil EXAFS Spectra in K-edge at various temperatures.
	}
	\label{fig:cutoff}
\end{figure}
\end{minipage}
\hfill
\begin{minipage}[t]{.46\linewidth}
\vspace{-0.3in}
\begin{figure}[H]
	\centering
	\includegraphics[trim=10 10 10 10,clip,width=\linewidth]{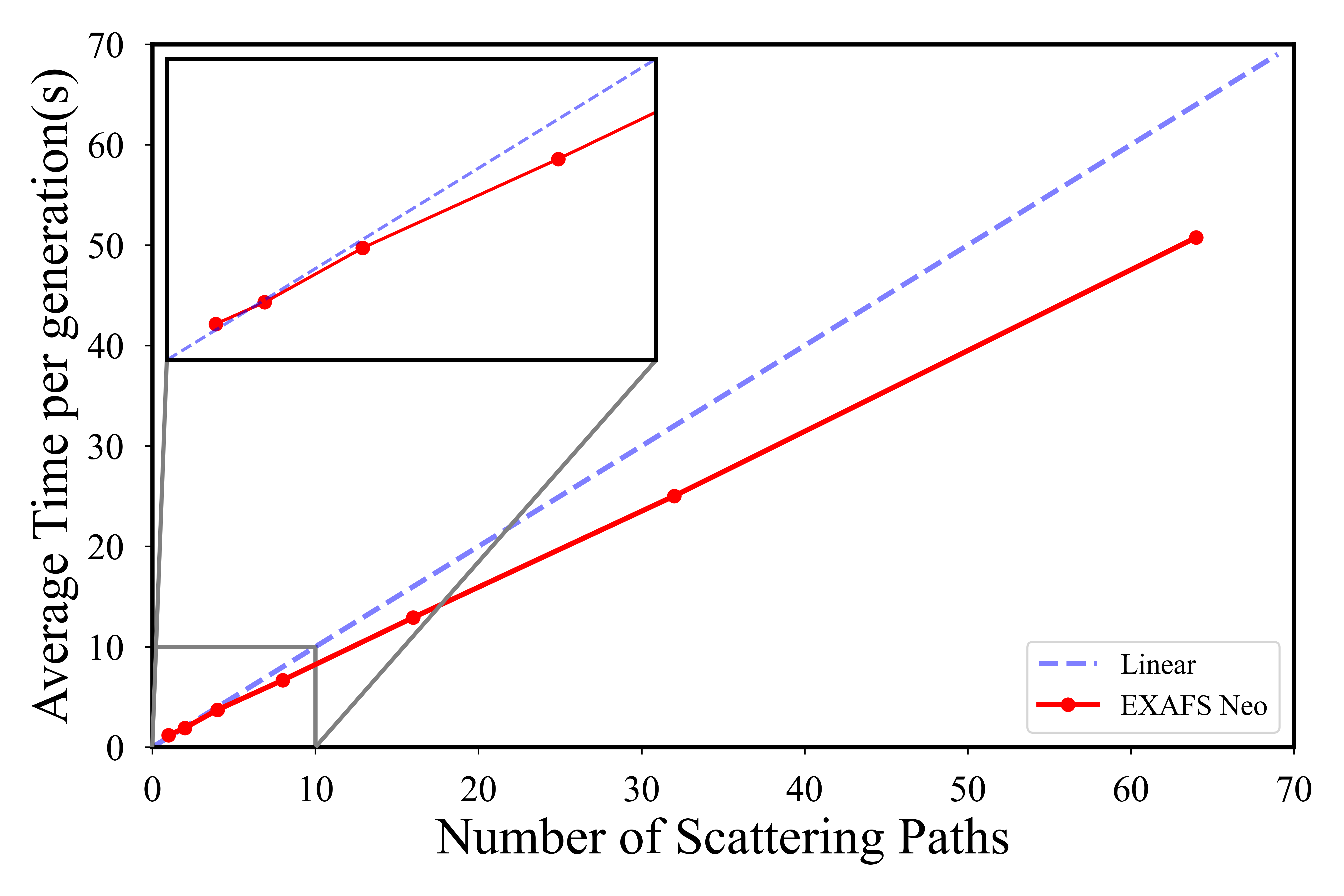}
	\caption{The complexity increases as the number of scattering paths increases.}
	\label{fig:TPG}
\end{figure}
\end{minipage}

\section{Conclusion}
\par We have developed a GA based software with the capability of efficiently performing automated materials characterization analysis of both complex X-ray spectra and nanoindentation data. 
We provided multiple crossover and mutation options in the code from which users can choose to optimize the analysis for their specific materials systems under study. We have extensively tested our software under various synthetic and experimental conditions. Our results demonstrated good scores of fitting without human inputs. We also tested the code with various cut off percentage to obtain the best scattering paths set. We note that caution must be exercised in selecting the data set over which to perform the cut off analysis. It must be representative of the entire collection of data to be analyzed for the results to be meaningful. The extensibility of our code base is a major advantage. Adding new analysis techniques does not require any new debugging of the core module of the GA code. 

\section{Acknowledgments}

\par This work was supported by author ML's startup fund. It is also supported through Idaho National Laboratory (INL) Laboratory Directed Research and Development (LDRD) Program under DOE Idaho Operations Office Contract DE-AC07-05ID14517. This research made use of the resources of the High Performance Computing Center at INL, which is supported by the Office of Nuclear Energy of the U.S. Department of Energy and the Nuclear Science User Facilities under Contract No. DE-AC07-05ID14517. 

\tiny
\printbibliography

@article{Terry2021,
title = {Analysis of extended X-ray absorption fine structure (EXAFS) data using artificial intelligence techniques},
journal = {Applied Surface Science},
volume = {547},
pages = {149059},
year = {2021},
issn = {0169-4332},
%doi = {https://doi.org/10.1016/j.apsusc.2021.149059},
url = {https://www.sciencedirect.com/science/article/pii/S0169433221001355},
author = {Jeff Terry and Miu Lun Lau and Jiateng Sun and et al.},
keywords = {Extended X-ray Absorption Fine Structure (EXAFS), Genetic algorithm, Synchrotron radiation, Machine learning, Artificial intelligence}
}

@inproceedings{Blaiszik2019,
    author = {Blaiszik, Ben and Chard, Kyle and Chard, Ryan and Foster, Ian and Ward, Logan},
    year = {2019},
    month = {01},
    pages = {020003},
    title = {Data automation at light sources},
    volume = {2054},
    journal = {AIP Conference Proceedings}
}

@article{Rehr2010,
	year = {2010},
	publisher = {Royal Society of Chemistry ({RSC})},
	volume = {12},
	number = {21},
	pages = {5503},
	author = {John J. Rehr and Joshua J. Kas and Fernando D. Vila and Micah P. Prange and Kevin Jorissen},
	title = {Parameter-free calculations of X-ray spectra with {FEFF}9},
	journal = {Physical Chemistry Chemical Physics}
}

@misc{Neo,
%	author = "{EXAFS Neo}",
	title = "{EXAFS Neo}",
	howpublished = {\url{https://github.com/laumiulun/EXAFS-Neo-Public}}
}

@article{sayers1971new,
	title={New Technique for investigating noncrystalline structures: Fourier analysis of the extended X-ray absorption fine structure},
	author={Sayers, Dale E and Stern, Edward A and Lytle, Farrel W},
	journal={Phys. Rev. Lett.},
	volume={27},
	number={18},
	pages={1204},
	year={1971},
	publisher={APS}
}

@inproceedings{newville2013larch,
	title={Larch: an analysis package for XAFS and related spectroscopies},
	author={Newville, Matthew},
	booktitle={J Phys Conf Ser},
	volume={430},
	pages={012007},
	year={2013}
}

@article{Ravel2005,
	year = {2005},
	month = jun,
	publisher = {International Union of Crystallography ({IUCr})},
	volume = {12},
	number = {4},
	pages = {537--541},
	author = {B. Ravel and M. Newville},
	title = {{ATHENA}, {ARTEMIS}, {HEPHAESTUS}: data analysis for X-ray absorption spectroscopy {using IFEFFIT}},
	journal = {Journal of Synchrotron Radiation}
}

@misc{WinXAS,
	title = "{WinXAS v3.2}",
	howpublished = {\url{http://www.winxas.de}},
	note = {Retrieved: 2020-10-30}
}

@article{bunker2005new,
	title={New methods for EXAFS analysis in structural genomics},
	author={Bunker, Grant and Dimakis, Nicholas and Khelashvili, Gocha},
	journal={Journal of Synchrotron Radiation},
	volume={12},
	number={1},
	pages={53--56},
	year={2005},
	publisher={International Union of Crystallography}
}

@article{holland1992genetic,
	title={Genetic algorithms},
	author={Holland, John H},
	journal={Scientific American},
	volume={267},
	number={1},
	pages={66--73},
	year={1992},
	publisher={JSTOR}
}

@article{goldberg1988genetic,
  title={Genetic Algorithms and Machine Learning},
  author={Goldberg, David E and Holland, John H},
  journal={Machine Learning},
  volume={3},
  number={2},
  pages={95--99},
  year={1988},
  publisher={Springer}
}

@book{eiben2003introduction,
	title={Introduction to Evolutionary Computing},
	author={Eiben, Agoston E and Smith, James E and others},
	volume={53},
	year={2003},
	publisher={Springer}
}

@article{bevington1993data,
    title={Data Reduction and error analysis for the physical sciences},
    author={Bevington, Philip R and Robinson, D Keith and Blair, J Morris and Mallinckrodt, A John and McKay, Susan},
    journal={Computers in Physics},
    volume={7},
    number={4},
    pages={415--416},
    year={1993},
    publisher={American Institute of Physics}
}

@inproceedings{rechenberg1994evolutionsstrategie,
  title={Evolutionsstrategie'94},
  author={I. Rechenberg},
  booktitle={Werkstatt Bionik und Evolutionstechnik},
  year={1994}
}

@article{chib1995understanding,
	title={Understanding the metropolis-hastings algorithm},
	author={Chib, Siddhartha and Greenberg, Edward},
	journal={The American Statistician},
	volume={49},
	number={4},
	pages={327--335},
	year={1995}
}

@book{meyer1965introductory,
	title={Introductory probability and statistical applications},
	author={Meyer, Paul L},
	year={1965},
	publisher={Oxford and IBH Publishing}
}

@phdthesis{redhouse2017uncertainty,
	title={Uncertainty Quantification of a Genetic Algorithm for Neutron Energy Spectrum Adjustment},
	author={Redhouse, Danielle R},
	year={2017}
}

@article{major2020,
    author = {Major,George H. and Avval,Tahereh G. and Moeini,Behnam and et al.},
    title = {Assessment of the frequency and nature of erroneous X-ray photoelectron spectroscopy analyses in the scientific literature},
    journal = {Journal of Vacuum Science \& Technology A},
    volume = {38},
    number = {6},
    pages = {061204},
    year = {2020}
}

@book{Bunker2009,
    year = {2009},
    publisher = {Cambridge University Press},
    author = {Grant Bunker},
    title = {Introduction to {XAFS}}
}

\end{document}